\documentclass[runningheads]{llncs}

\usepackage[T1]{fontenc}
\usepackage{xcolor}
\usepackage{bm}
\usepackage{amsmath,amssymb,amsfonts}
\usepackage[mathscr]{euscript}
\usepackage[ruled,linesnumbered,vlined,shortend]{algorithm2e}
\usepackage{algorithmicx}
\usepackage{graphicx}
\usepackage{stackengine}
\usepackage{booktabs}
\usepackage{subfigure}
\usepackage{multirow}
\usepackage{tabularx}
\usepackage{tablefootnote}
\usepackage{threeparttable}
\usepackage{caption}
\usepackage{mdwlist}
\usepackage{makecell}
\usepackage{enumitem}
\usepackage{diagbox}
\usepackage{hyperref}

\usepackage[capitalise,nameinlink]{cleveref}
\crefname{section}{Sec.}{Sec.}
\Crefname{section}{Section}{Sections}
\crefname{listing}{List.}{List.}
\crefname{listing}{Listing}{Listings}
\Crefname{listing}{Listing}{Listings}
\crefname{lstlisting}{Listing}{Listings}
\Crefname{lstlisting}{Listing}{Listings}

    \makeatletter
\def\@fnsymbol#1{\ensuremath{\ifcase#1\or \dagger\or *\or
   \mathsection\or \mathparagraph\or \|\or **\or \dagger\dagger
   \or \ddagger\ddagger \else\@ctrerr\fi}}
    \makeatother

\newcommand{\T}[1]{\bm{\mathscr{#1}}}

\newcommand{\mat}[1]{\mathbf{#1}}
\newcommand{\vect}[1]{\mathbf{#1}}

\SetKwInput{KwInput}{Input}
\SetKwInput{KwOutput}{Output}

\newcommand{\blue}[1]{{\color{black} #1}}

\newcommand{\hide}[1]{}
\newcommand{\method}[0]{\textsc{STAFF}\xspace}
\newcommand{\fullmethod}[0]{Sparse Tensor Augmentation For Fairness}
\newcommand{\cpd}[0]{CPD\xspace}

\newcommand{\costco}[0]{CoSTCo\xspace}

\newcommand{\dain}[0]{DAIN\xspace}
\newcommand{\random}[0]{RANDOM\xspace}

\newcommand{\lastfm}[0]{{LastFM}\xspace}

\newcommand{\oulad}[0]{{OULAD}\xspace}

\newcommand{\chicago}[0]{{Chicago-Crime}\xspace}

\begin{document}
\title{Improving Group Fairness in Tensor Completion via Imbalance Mitigating Entity Augmentation}
\titlerunning{\fullmethod}
\authorrunning{Dawon Ahn, Jun-Gi Jang, and Evangelos E. Papalexakis}

\newcommand{\repeatthanks}{\textsuperscript{\thefootnote}}
\author{Dawon Ahn\inst{1}\thanks{Corresponding author.}\thanks{These authors contributed equally to this work.} \orcidID{0009-0003-9268-4058} \and
Jun-Gi Jang\inst{2}\repeatthanks\orcidID{0000-0001-8328-3920} \and
Evangelos E. Papalexakis\inst{1}\orcidID{0000-0002-3411-8483}}
\institute{Department of Computer Science and Engineering, University of California, Riverside, Riverside, CA, USA
\and 
Siebel School of Computing and Data Science,  University of Illinois Urbana-Champaign, Urbana, IL, USA \\
\email{dahn017@ucr.edu, jungi@illinois.edu, epapalex@cs.ucr.edu}
}

\maketitle              
\begin{abstract}
Group fairness is important to consider in tensor decomposition 
to prevent discrimination based on social grounds such as gender or age.
Although few works have studied group fairness in tensor decomposition, 
they suffer from performance degradation. 
To address this, we propose \method (\fullmethod) to improve group fairness by minimizing the gap in completion errors of different groups while reducing the overall tensor completion error.
Our main idea is to augment a tensor with augmented entities including sufficient observed entries to mitigate imbalance and group bias in the sparse tensor.
We evaluate \method on tensor completion with various datasets under conventional and deep learning-based tensor models.
\method consistently shows the best trade-off between completion error and group fairness;
at most, it yields $36 \%$ lower MSE and $59 \%$ lower MADE than the second-best baseline.

\end{abstract}

\section{Introduction}
\label{sec:introduction}
Given an incomplete tensor where one of its modes includes sensitive attributes, 
how can we accurately complete the tensor while reducing the gap between errors of each group based on these attributes?
Tensors are a natural way to represent multi-aspect data.
For example, a crime dataset can be represented
	as a third-order tensor with (time, location, crime type) modes,
	where each value indicates the number of incidents.
Tensor decomposition is a fundamental model for discovering latent patterns in tensors
	by representing them as factor matrices~\cite{kolda2009tensor}.
Tensor models have proven effective in tensor completion,
	a task of predicting missing entries in a tensor based on observed entries,
	with various applications in tensor streams~\cite{jang2023static,qiu2025tucket}, 
	social network~\cite{sun2009multivis,papalexakis2014spotting},
	healthcare~\cite{afshar2020taste,ho2014marble}, and
	education~\cite{wang2021stretch,wang2021knowledge}. 

Recently, group fairness has increased its importance in the development of data mining algorithms
	to prevent discrimination of those algorithms against groups
	identified by sensitive attributes or social grounds such as gender, race, or age~\cite{barocas2023fairness,chen2023improving,kim2021learning,mehrabi2021survey,sun2023fair,yao2017beyond,zhu2018fairness}.
Group unfairness stems from historical bias in input data,
such as discriminatory customs or unequal resource allocation among groups~\cite{barocas2023fairness,mehrabi2021survey}.
This bias often results in imbalanced data collection between groups,
making those algorithms produce discriminatory outcomes~\cite{chen2023improving}.
A similar challenge occurs in tensor completion for sparse tensors.
When a tensor exhibits an imbalanced number of observed entries among groups, 
a gap of tensor completion error among groups increases, as shown in \Cref{fig:cpd_costco}.
Tensor models predicts missing entries accurately for the majority having more data while predicts missing entry less accurately for the minority having less data.
This outcome results in discrimination against the minority group, reinforcing the existing bias.
Thus, our fairness goal is to reduce the gap in completion error of different groups when those groups have different number of observed entries.

However, no studies have explored group fairness in tensor completion with regard to error differences. 
Few works~\cite{kim2021learning,zhu2018fairness} aim to achieve a demographic parity as a group fairness goal, which makes tensor models independent to sensitive attributes.
This fairness concept is only appropriate when a given task is unrelated to the sensitive attributes~\cite{yao2017beyond}, 
otherwise achieving this fairness concept in the model will hurt the performance, decreasing the utility of the model.
Similar to recent methods~\cite{chen2023improving,sun2023fair} that improved both performance and group fairness by mitigating data bias, we can employ a tensor augmentation method~\cite{oh2021influence} that helps alleviate data imbalance in an incomplete tensor by generating tensor entries.
However, this method is limited in achieving group fairness since it tends to generate better quality for the majority while it does not for the minority.

In this paper, we propose \method (\fullmethod) to improve group fairness in tensor completion.
The main idea is to augment sensitive entities with insufficient observed entries,
enhancing their representation and reducing the completion error gap between groups.
To generate an augmented entity, we leverage neighbors of the original entity
with a fairness-aware graph that identifies neighbors with a context similarity and sensitive attribute.
We then generate observed entries for the augmented entity leveraging data of the original entity and the identified neighbors.
Finally, we regularize the original entities with the augmented ones to incorporate the augmentation effect during tensor decomposition.
Our contribution is as follows.
\begin{itemize}[topsep=1pt]
        \item \textbf{Fairness-aware augmentation and regularization.} We propose a novel approach that improves group fairness by mitigating imbalance and group bias in a sparse tensor.
        \item \textbf{Model-agnostic approach.} \method improves group fairness and completion performance in both traditional and deep learning-based tensor models.
	\item \textbf{Experiments.} We experimentally show that \method outperforms baselines regarding completion error and group fairness across models and datasets.
\end{itemize}
The source code and datasets are available at \href{https://github.com/dawonahn/STAFF/}{https://github.com/dawonahn/STAFF/}.

\section{Preliminaries \& Related Work}
\label{sec:prelim}
We explain tensor decomposition models and discuss related works on augmentation and fairness.
We then formally define a problem definition.
\vspace{-5mm}
\subsubsection{Tensor Decomposition.} \label{sec:prelim:tensor}
Conventional tensor models such as Tucker or CANDECOMP/PARAFAC (CP) decomposition 
have been developed for tensor completion across various applications~\cite{kolda2009tensor,papalexakis2016tensors}.
To improve the capability to capture complex patterns, 
several deep learning-based tensor models have been proposed~\cite{ahn2024neural,liu2019costco}.
We select most widely used tensor models, CP decomposition and CoSTCo~\cite{liu2019costco},
as base models in our evaluation.
We define a general form of tensor decomposition.
Given an $N$-order tensor  $\T{X} \in \mathbb{R}^{I_1 \times I_2 \times \cdots \times I_N}$ and rank $R$, 
tensor decomposition approximates the tensor with factor matrices $\{ \mat{U}^{(n)} \in \mathbb{R}^{I_n \times R} \vert n = 1, \cdots, N \}$ and other training parameters $\theta$
that minimize the following loss function:
\begin{equation}  \label{eq:tf}
\small
L = \sum_{\alpha = (i_1, i_2, \cdots, i_N) \in \Omega}{ \left({x}_{\alpha} - \tilde{{x}}_{\alpha}\right)^{2}}
\text{ with } \tilde{x}_{\alpha} = f(\theta, \{\vect{u}^{(1)}_{i_1}, \vect{u}^{(2)}_{i_2}, \cdots, \vect{u}^{(N)}_{i_N} \}).
\end{equation}
Note that
$\Omega$ indicates a set of indices $\alpha$ for tensor entries
and $\tilde{{x}}_{\alpha}$ is the reconstructed entry corresponding to index $\alpha$.
Here, $\vect{u}^{(n)}_{i_n}$ is the $i_n$th row of the $n$th factor matrix, representing the $i_n$th entity. 
\Cref{eq:tf} becomes CP decomposition when factor matrices are used as parameters and $f$ is defined as:
\begin{equation}  \label{eq:cpd}
\small
\tilde{{x}}_{\alpha} = \sum_{r=1}^{R} u^{(1)}_{i_1r} u^{(2)}_{i_2r} \cdots u^{(N)}_{i_Nr},
\end{equation}
where $u^{(n)}_{i_nr}$ is the $r$th element of $\vect{u}^{(n)}_{i_n}$.
\Cref{eq:tf} becomes CoSTCo if $f$ is defined as:
\begin{equation}  \label{eq:costco}
\small
\tilde{{x}}_{\alpha} = MLP(Conv(Conv(\T{Z}_{\alpha}, \Theta_1), \Theta_2), \Theta_3),
\end{equation}
where $\T{Z}_{\alpha} = [\vect{u}^{(1)}_{i_1}; \vect{u}^{(2)}_{i_2}; \cdots;\vect{u}^{(N)}_{i_N}] \in \mathbb{R}^{1 \times N \times R}$ a concatenation of row factors and 
$Conv$ and $MLP$ denote the operation of 1-D Convolutional Neural Network and Multi-Layer Perceptron, respectively,
and $\Theta_1, \Theta_2,$ and $\Theta_3$ indicates learnable parameters in these neural networks.
To complete missing entries, 
these tensor models reconstruct missing entries with factor matrices and learnable parameters trained on observed entries.
The tensor completion error is measured as the difference between reconstructed and missing entries as \Cref{eq:tf}.
\vspace{-5mm}
\subsubsection{Augmentation.} \label{sec:prelim:augmentation}
Augmentation techniques aim to enhance the quantity, quality, and diversity of training data by modifying the data slightly while preserving the semantics of the original data~\cite{mumuni2022data}.
In tensor completion, Oh et al.~\cite{oh2021influence} proposed \dain, 
a sparse tensor augmentation method that augments missing entries in the original tensor.
It calculates the influence score of entities with neural networks 
that have a high impact on the completion accuracy.
It selects indices of missing entries by sampling entries based on the importance score of entities
and predicts values of the sampled entries with tensor decomposition.
However, this approach often tends to augment entries of entities having more data,
which can be less accurate for entities with less data.
Also, suboptimal augmentation for the missing entries can introduce noise 
since the entries of an original tensor can be overlapped with test entries.
In contrast, we augment sensitive entities having insufficient entries to improve their representation, thereby reducing the completion error gap of groups.

\vspace{-5mm}
\subsubsection{Group Fairness} \label{sec:prelim:fairness}
Measuring group fairness is challenging as there is no universal definition;
metrics depend on the context of real-world applications~\cite{barocas2023fairness,mehrabi2021survey}.
Demographic Parity (DP)~\cite{dwork2012fairness} and Equalized Odds (EO)~\cite{hardt2016equality} are widely known metrics to measure group fairness in classification tasks;
DP ensures that model outcomes are independent of sensitive attributes, ignoring true labels, while EO ensures predictive performance by requiring that the rates of true positives and false positives are equal between groups.
Recent works~\cite{chan2024group,han2024hypergraph,zhao2023fair} have focused on balancing group fairness and utility.
Similarly in tensor completion, the difference in reconstruction error between groups is important for ensuring group fairness.
When there is a significant error gap between groups, 
the inaccurate reconstruction of the minority reflects they are not well represented in tensor models.
Similar to the EO metric in the classification tasks,
we define Mean Absolute Difference of Error (MADE) for group fairness in tensor completion, measuring the mean difference of reconstruction errors of \blue{missing entries} between groups $g_1$ and $g_2$, as follows.
 \begin{equation} \label{eqref:made}
\small
 \text{MADE: } \left| \frac{1}{|\Omega_{g_1}|}\sum_{\alpha_1 \in \Omega_{g_1}} |x_{\alpha_1} - \tilde{x}_{\alpha_1} | -
 		 \frac{1}{|\Omega_{g_2}|}\sum_{\alpha_2 \in \Omega_{g_2}}{|x_{\alpha_2} - \tilde{x}_{\alpha_2}|} \right|
 \end{equation}
\subsubsection{Fairness-Aware Tensor Models} \label{sec:prelim:fairness}
There are few studies addressing group fairness in matrix and tensor decomposition models~\cite{kassab2024towards,kim2021learning,yao2017beyond,zhu2018fairness}.
Yao et al.~\cite{yao2017beyond} studied various types of fair constraints based on reconstruction error for collaborative filtering.
Kassab et al.~\cite{kassab2024towards} propose nonnegative matrix factorization minimizing the maximum reconstruction loss among groups.
Recent works~\cite{kim2021learning,zhu2018fairness} proposed fairness-aware tensor models improving group fairness by making the model independent of sensitive attributes. 
Zhu et al.~\cite{zhu2018fairness} proposed FATR for recommendation by removing sensitive attribute-related latent patterns in factor matrices.
Kim et al.~\cite{kim2021learning} proposed a Kernel Hilbert Schmidt Independence Criterion (KHSIC) based regularization making sensitive attributes independent of a factor matrix.
However, these approaches can increase completion error since those attributes are often important in completion~\cite{barocas2023fairness,mehrabi2021survey}. 
We improve group fairness with regard to tensor completion errors with data perspective while maintaining the performance.

\subsection{Problem Definition} \label{sec:prelim:pb}
We define terms and a problem in a group fairness setting as follows.
The mode $s$ associated with sensitive attributes is referred to as the sensitive mode, and 
its entities and the corresponding factor matrix are referred to as sensitive entities and the sensitive factor matrix, respectively.
Non-sensitive factor matrices and a sensitive factor matrix
are denoted as $\{ \mat{U}^{(n)} \in \mathbb{R}^{I_n \times R} \vert n \neq s \}$ and $\mat{U}^{(s)} \in \mathbb{R}^{I_s \times R}$, respectively.
The problem definition is as follows:
\textbf{\textit{given}} (1) an $N$th order sparse tensor with a sensitive mode $s$, and (2) sensitive attribute matrix,
 \textbf{\textit{find}} a tensor decomposition that minimizes MADE (\Cref{eqref:made}) and completion error (\Cref{eq:tf}).
\section{Proposed Method}
\label{sec:method}
\begin{figure*}[!tp]
	\centering
	\includegraphics[width=\linewidth]{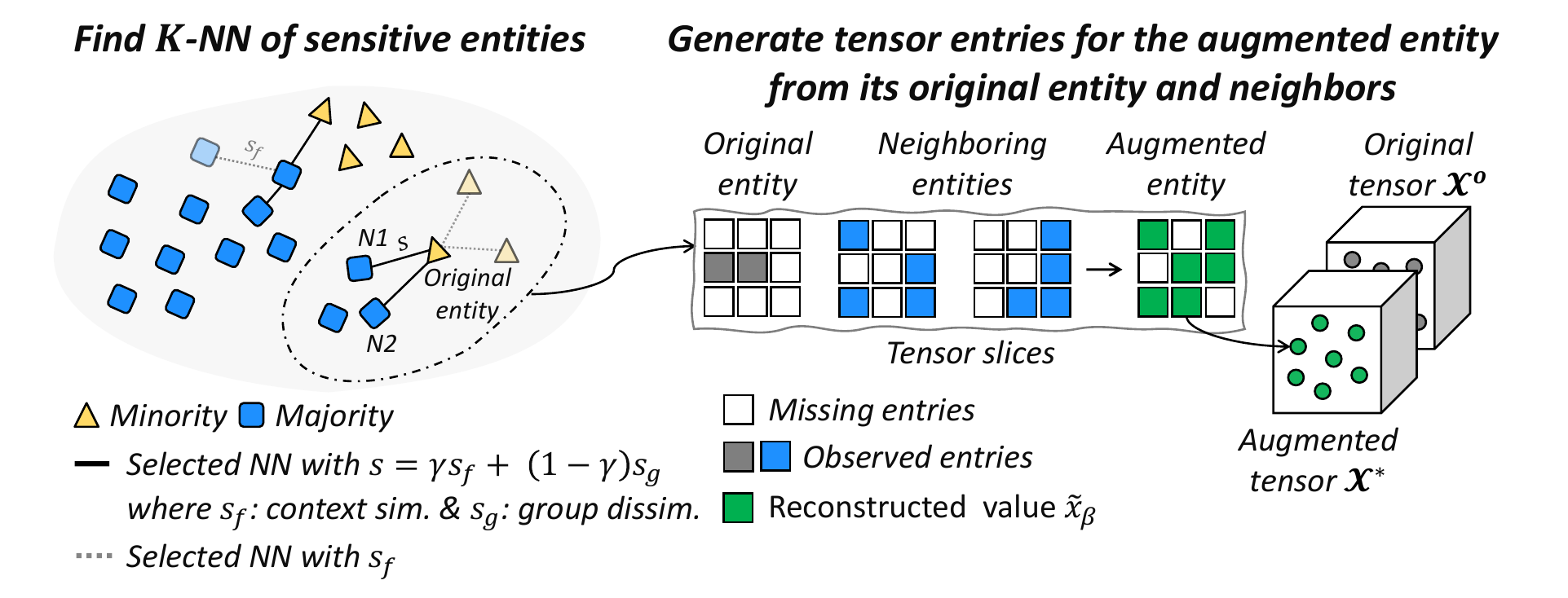}
    \caption{ \label{fig:overview} 
        The main idea of \method.
 }
\end{figure*}

We propose \method, to reduce the gap between errors of each group caused by data imbalance in tensor completion.
There are two main challenges as follows.
\begin{itemize}[topsep=2.5pt]
	\item \textbf{Imbalance bias:} 
    the minority lacking sufficient observed entries has a higher completion error than the majority having enough entries.
	\item \textbf{Group bias:} tensor decomposition models favor the majority over the minority
			when the interaction of the majority is critical in minimizing the overall completion error.
\end{itemize}
The following ideas reduce MADE in Equation~\eqref{eqref:made}, the difference in reconstruction error between groups, by addressing the challenges:
\begin{itemize}[topsep=2.5pt]
	\item \textbf{$K$-NN graph guided augmentation and its associated regularization}
            reduce the completion errors of sensitive entities with a limited number of tensor entries, particularly those belonging to the minority group.
	\item \textbf{Group fairness-aware $K$-NN graph construction}
            mitigates favoritism for the majority group by encouraging the selection of cross-group neighbors.
\end{itemize}
We generate an augmented view $\T{X}^{*}$ where it has the same size as a given tensor $\T{X}^{o}$ and sensitive entities are augmented.
To decide which entries to fill in $\T{X}^{*}$, we construct a group fairness-aware $K$-NN graph, which selects neighbors by balancing context similarity and group fairness.
Augmented entities have more tensor entries compared to the ones that their corresponding entities in the original tensor have.
We generate synthetic entries for each augmented entity using the entries from neighboring entities of its corresponding original entity. 
\Cref{fig:overview} illustrates how to construct a fairness-aware graph and generate tensor entries for the augmented entity based on the graph.
We then incorporate regularization into a tensor decomposition model to minimize the difference between the representations of the original entities and those of augmented entities.
\vspace{-3mm}
\subsection{Fairness-Aware Augmentation for Sensitive Entity } \label{sec:method:augmentation}
The goal is to generate an augmented tensor $\T{X}^{*}$ for sensitive entities.
We create augmented entities that are similar to their original entities but include an abundant number of tensor entries.
Selecting which entries to augment in a tensor slice is important for ensuring the augmented entities similar to the original entities. 
For example, if tensor entries in a slice are randomly augmented, the resulting factors of the augmented entities significantly differ from those of the original entities.
Thus, we leverage neighbors sharing similar patterns with the original entity to generate entries.
\vspace{-3mm}
\subsubsection{Group Fairness-Aware Graph} \label{sec:method:fairgraph}
To find neighbors, we use a similarity score $s$ based on a context feature and a sensitive attribute.
We use a sensitive factor matrix $\mat{U^{(s)}} \in \mathbb{R}^{I_s \times R}$ as the context feature, obtained by decomposing the original tensor with \Cref{eq:tf}.
Also, we create a one-hot encoded sensitive feature $\mat{F} \in \mathbb{R}^{I_s \times M} $ with the sensitive attribute, indicating entities' group membership,
e.g., for gender with two categories, female and male, the $i_s$th sensitive feature is $\vect{f}_{i_s} = [0 \text{ } 1]$ or $[1 \text{ } 0]$, 
We construct a $K$-NN graph $G$ by selecting $K$ neighboring entities based on the similarity score $s$ that combines feature similarity $s_f$ and group dissimilarity $s_g$ for every $i, j \in I_s$ as:
\begin{equation}  \label{eq:staff:graph}
s(i, j) = \gamma s_f (i, j) + ( 1 - \gamma) s_g (i,j) \\
= 
\gamma \frac{\vect{u}^{(s)}_i \cdot \vect{u}^{(s)}_j}{\lVert \vect{u}^{(s)}_i \rVert  \lVert \vect{u}^{(s)}_{j} \rVert} 
+ (1 - \gamma)(1 - \frac{\vect{f}_i \cdot \vect{f}_j}{\lVert \vect{f}_i \rVert \lVert  \vect{f}_{j} \rVert}), 
\end{equation}
where $\gamma$ balances the weight of $s_f$ and $s_g$.
Here, $s_f$ encourages the selection of neighbors with similar factors, while $s_g$ encourages the selection of neighbors from different groups.
The score $s$ allows for selecting the most similar neighbors both within the same group and across different groups. 
\vspace{-3mm}
\subsubsection{Tensor Entry Augmentation} \label{sec:method:augmentation}
The augmented tensor $\T{X}^{*}$ needs to retain the original entity's patterns while having diverse contextual variations for fairness.
To achieve it, we selectively generate tensor entries of the augmented tensor $\T{X}^{*}$ using the $K$-NN graph $G$.
Instead of using all observed tensor entries from the original tensor, 
we randomly sample entries from the original entity and their neighbors, respectively.
First, we randomly sample $P$ number of tensor entries from the original entity  $\vect{u}_{i^{o}_{s}}$, denoted as, $\tau_{O} = \{ (\alpha, x_{\alpha}) \vert \alpha  = (i_1, \cdots, i_{s^o}, \cdots, i_N) \in \Omega_{i_s = i^{o}_{s}} \}$.
This set of entries preserves the characteristics of the original entity, making the augmented entity similar to its original entity.
We then generate tensor entries by randomly sampling $Q$ number of interactions from $K$ neighbor entities $N(i^{o}_{s}) = \{i^{o}_{s_k} | k \in K \}$,
denoted as $\tau_{N} = \{(\beta, x_{\beta}) \vert \beta = (i_1, \cdots, i^{o}_{s_k}, \cdots, i_N) \in \Omega_{i^{o}_{s_k} \in N(i^{o}_s)} \}$.
Even though the sampled interactions increase the volume of tensor entries,
	their values do not accurately represent the original entity $\vect{u}_{i^{o}_{s}}$.
Thus we replace $x_{\beta}$ as $\tilde{x}_{\beta}$ which is a predicted value \blue{with the averaged row factors of original sensitive entity and its neighbors} and non-sensitive mode factors with a given tensor model as following:
\begin{equation}
\tilde{x}_{\beta} = f(\theta, [\vect{u}_{i_1}, \cdots, \frac{1}{(K+1)}(\vect{u}_{i^{o}_s} + \sum_{k=1}^{K}\vect{u}_{i^{o}_{s_{k}}}), \cdots, \vect{u}_{i_N}]).
\end{equation}
If we predict ${x}_{\beta}$ with only $\vect{u}_{i^o_{s}}$,
the predicted value will be poor for the minority, 
since their factor vectors are poorly trained with insufficient entries.-
If we predict ${x}_{\beta}$ with only its neighbors, the predicted values can 
 be significantly different from the original entity when neighbors have different characteristics.
Finally, an augmented entity $\vect{u}_{i^{*}_{s}}$ corresponding to the original entity $\vect{u}_{i^{o}_s}$ has $P + Q$ number of augmented observed entries in \blue{the augmented tensor $\T{X}_{i^{*}_s} \in \mathbb{R}^{I_1 \times \cdots \times I_{s-1} \times I_{s+1} \times \cdots \times I_N}$.}
\vspace{-3mm}
\subsection{Fairness \& Imbalance-Aware Tensor Decomposition}  \label{sec:method:staff}
We utilize augmented entities to reduce the gap of errors of groups by enhancing tensor completion of both the minority and the majority.
Given an input tensor $\T{X} = [\T{X}^{o}; \T{X}^{*}]$ and a rank $R$, 
\method decomposes the tensor into 
non-sensitive factor matrices $\{ \mat{U}^{(n)} \in \mathbb{R}^{I_n \times R} \vert n \neq s \}$ and a sensitive factor matrix $\mat{U}^{(s)} \in \mathbb{R}^{(I_s^{o} + I_s^{*})\times R}$ with other training parameters $\Theta$,
by minimizing the following loss:
\begin{equation}  \label{eq:staff:ssl}
\small
L = \sum_{\alpha \in \Omega}{ \left({x}_{\alpha} - \tilde{{x}}_{\alpha}\right)^{2}}
+ \lambda_r \sum_{n=1}^{N} \lVert \mat{U}^{(n)}\rVert^{2}_{F}
+ \lambda_{g} \sum_{i_s=1}^{I_s} \lVert \vect{u}^{(s)}_{{i^{o}_s}} - \vect{u}^{(s)}_{{i^{*}_s}} \rVert^{2},
\end{equation}
where $\Omega$ includes the observed entries of the original and augmented tensors, and 
$\lambda_r$ and $\lambda_f$ are coefficients to adjust regularizations.
The second term indicates a L2 regularization, and the third term $\sum_{i_s=1}^{I_s} \lVert\vect{u}^{(s)}_{{i^{o}_s}} - \vect{u}^{(s)}_{{i^{*}_s}} \rVert^{2}$ 
regularizes the factor matrix of the original entities with one of their corresponding augmented entities. 
This regularization incorporates the augmentation effect into the factor matrix effectively and 
enhances the representation of sensitive entities 
to reduce the disparity of completion error between groups.

\vspace{-3mm}
\section{Experimental Evaluation}
\label{sec:experiment}
\vspace{-3mm}
In this section, our aim is to answer the following questions.
\begin{itemize*}
	\item [\textbf{Q1}] {
		\textbf{Fairness \& Performance (\Cref{sec:exp:performance}).}
		\blue{How fair and accurate does \method complete missing entries in group imbalanced tensors?}
	}
	\item [\textbf{Q2}] {
		\textbf{Hyper-Parameter Study (\Cref{sec:exp:hparams}).}
		How the hyper-parameters $\lambda_f$ and $\gamma$ affect fairness and accuracy in \method?
	}
\end{itemize*}
\vspace{-3mm}
\subsection{Setting}
For experiments, we utilize a machine equipped with an AMD EPYC 7313 16-Core Processor and an NVIDIA RTX A4000.

\noindent \textbf{Dataset.}
We use real-world tensors to evaluate group fairness and accuracy of the proposed method, summarized in \Cref{tab:dataset}.\lastfm is a music recommendation dataset represented as a third-mode tensor (user $\times$ artist $\times$ timestamp).
Each tensor entry is binary indicating whether user $i$ listens to the music of artist $j$ at the timestamp $k$.
We use the gender of the user as the sensitive attribute: female and male.
We sample 10$\%$ of artists and remove users having less than three interactions.
\oulad is a dataset of student interactions collected from the Open University Learning Platform represented as the third mode tensor (student $\times$ module $\times$ exam).
Each tensor entry is the score of the exam $k$ in the module by a student $i$.
	The sensitive attribute is the disability of the student: yes and no.
\chicago is a crime reports dataset in Chicago city represented as the third mode tensor (hour $\times$ area $\times$ crime type).
Each tensor entry denotes the number of crimes $k$ that occurred in community $j$ at hour $i$.
The sensitive attribute is a region of the community: north and south.
We split tensors into training, validation, and test datasets with a ratio of 8:1:1.
To examine whether the sparsity of the entries affects group fairness, 
we increase the sparsity of the minority by randomly sampling observed entries of the minority at rates of $10\%$, and $5\%$.
\begin{table*}[!t]
	\centering
	\small{
	\caption{ \label{tab:dataset}
		Summary of real-world tensors. 
		Bold text indicates a sensitive mode including a sensitive attribute.}
	\setlength{\tabcolsep}{1.5pt}
	\begin{tabular}{ l c c c c c c c}
		\toprule
		\textbf{Dataset} & \multicolumn{3}{c}{\textbf{Mode}} & \textbf{Nonzeros} & \textbf{Group} & \textbf{Majority} & \textbf{Minority} \\
		\midrule
		\multirow{2}{*}{\lastfm{}~\tnote{1}}  & \textbf{User} & Artist & Time & Interaction & \multirow{2}{*}{Gender} & Male & Female\\
		\cmidrule{2-5}\cmidrule{7-8}
		& 853 & 2,964 & 1,586 & 143,107 &  & 93,316 & 49,791 \\
		\midrule 
		\multirow{2}{*}{\oulad{}~\tnote{2}}  & \textbf{Student} & Module & Test & Score & \multirow{2}{*}{Disability} & No & Yes \\
		\cmidrule{2-5} \cmidrule{7-8}
		& 3,248 & 22 & 3 & 11,742 &  & 10,650 &  1,092 \\
		\midrule 
		\multirow{2}{*}{\chicago{}~\tnote{3}}  & Hour & \textbf{Area} & Crime & Count & \multirow{2}{*}{Location} & South & North \\
		\cmidrule{2-5} \cmidrule{7-8}
		& 24 & 77 & 32 & 42,097 &  & 23,723 & 18,374  \\
	 	\bottomrule
	\end{tabular}	
	}
\end{table*} 

\vspace{-1mm}
\noindent \textbf{Baselines.}
We use \cpd and \costco as base tensor models
for the proposed method and the following competitors.
Mean Absolute Difference of Reconstruction (MADR) indicates a tensor model with a MADR constraint defined as $
\left| \frac{1}{|\Omega_{g_1}|}\sum_{\alpha_1 \in \Omega_{g_1}} |\tilde{x}_{\alpha_1} | -
		 \frac{1}{|\Omega_{g_2}|}\sum_{\alpha_2 \in \Omega_{g_2}}{|\tilde{x}_{\alpha_2}|} \right|$.
MADE indicates a tensor model with a MADE constraint defined in \Cref{eqref:made}.
KHSIC~\cite{kim2021learning} is a tensor model with a KHSIC-based regularization.
\random~\cite{oh2021influence} is a tensor model trained with an augmentation, randomly augmenting missing entries with reconstructed values based on a tensor model.
DAIN~\cite{oh2021influence} is a tensor model trained with an augmentation, augmenting missing entries based on influences score.
Among fairness-aware tensor models~\cite{kim2021learning,zhu2018fairness}, we chose KHSIC as our baseline
since these models aim to achieve the same fairness concept, but KHSIC outperforms FATR. 

\noindent \textbf{Metric \& Training \& Hyper-Parameters} 
We use Mean Squared Error (MSE) to measure the completion error defined as:
$\frac{1}{|\Omega|}\sum_{\alpha \in \Omega}(x_{\alpha} - \tilde{x}_{\alpha})^2$.
We train tensor models with Adam~\cite{kingma2014adam} with a batch size of 1024.
We fix the rank as 10 and find optimal hyper-parameters: weight decay $\lambda_r$ terms and learning rate for all methods with datasets.
For \method, we pick $\lambda_f$ for the fairness-aware regularization from $1e-4$ to $1e+2$ with a log scale,
and pick $K$ to select neighbors from $\{3, 5, 7, 9, 11, 13, 15\}$.
We sample 30 entries from the original and neighboring entities to generate entries for the augmented entity.

\vspace{-3mm}
\subsection{Fairness \& Performance} \label{sec:exp:performance}
\begin{figure*}[!htp]
	\centering{
    \includegraphics[width=0.85\linewidth]{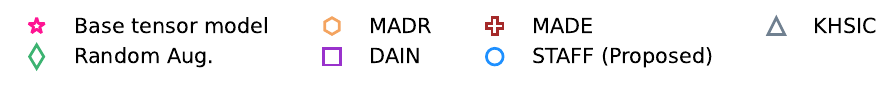}
    \subfigure[\cpd]{ \label{fig:cpd_costco:cpd}
     \includegraphics[width=0.76\linewidth]{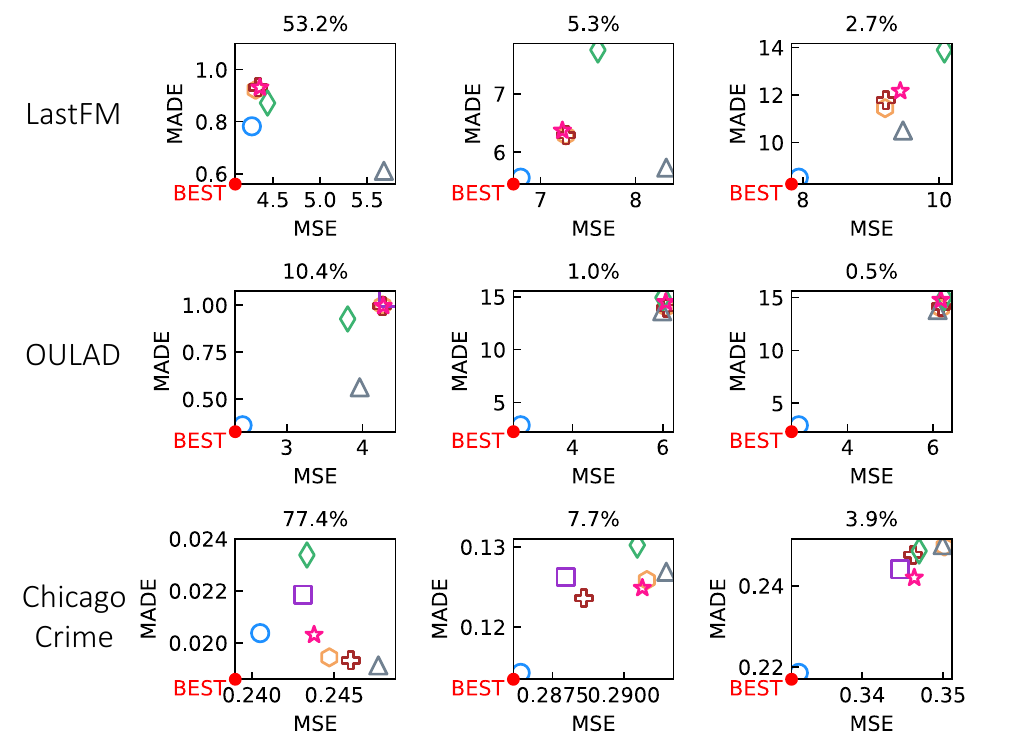}
    }
    \subfigure[\costco]{ \label{fig:cpd_costco:costco}
     \includegraphics[width=0.72\linewidth]{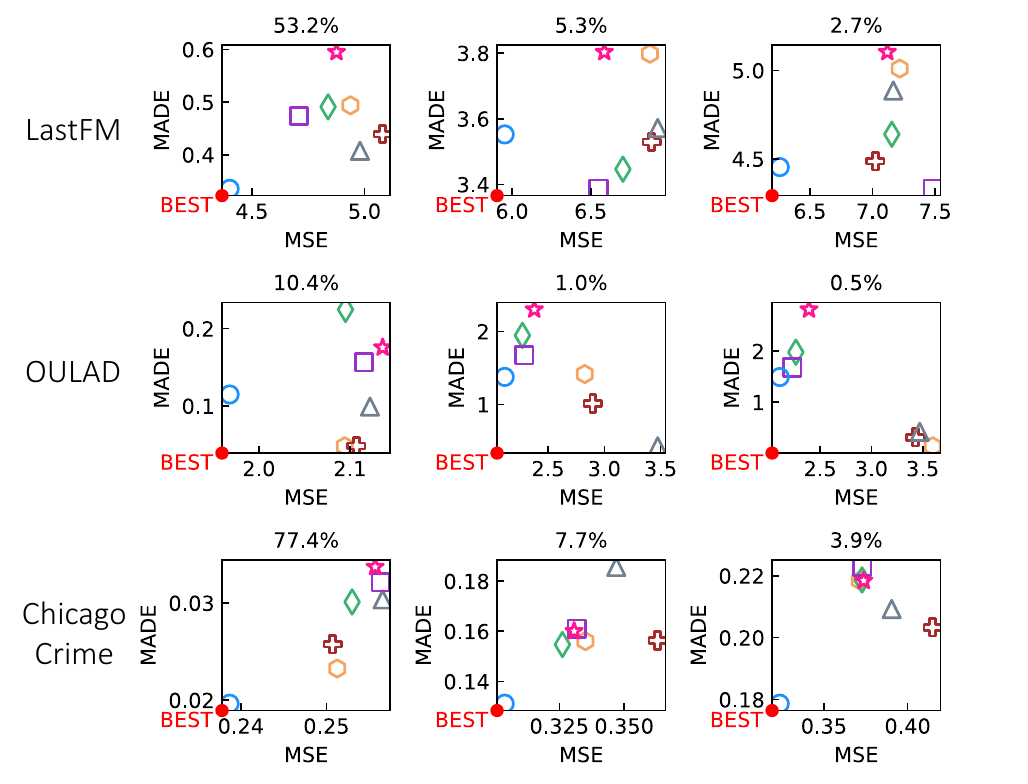}
    }
    \caption{\label{fig:cpd_costco}
    Trade off of MSE ($\times 10^{2}$) and MADE ($\times 10^{2}$).
    Each plot title indicates the sparsity of the minority compared to the majority, and the leftmost column is the original tensor.
    \method shows the best trade-off between MSE and MADE.}
 }
\end{figure*}

We evaluate \method compared to baselines in terms of MSE and MADE with two tensor models and three real-world datasets.
We vary the sparsity of the observed entries for the minority group to examine whether the sparsity is the source of unfairness which increases the gap of error between groups.
As shown in \Cref{fig:cpd_costco}, \method consistently achieves the best trade-off between MSE and MADE across all datasets and models.
\blue{With \cpd and an original \oulad dataset (first column), \method exhibits 
MSE of 0.0242 ($36 \% \downarrow$) and MADE of 0.0037 ($59 \% \downarrow$) while
MADE constraint, the second best method, shows MSE of 0.0380 and MADE of 0.0082.
With \costco and an original \lastfm dataset (first column), \method exhibits MSE of 0.047 ($6 \% \downarrow$) and MADE of 0.0084 ($29 \% \downarrow$) while
MADE constraint, the second best method, shows MSE of 0.0431 and MADE of 0.0092.
}
Note that, \method reduces MADE by decreasing errors for both the majority and the minority groups
while competitors achieve lower MADE by increasing errors of both groups.

\vspace{-3mm}
\subsection{Hyper-parameter Study} \label{sec:exp:hparams}
\begin{figure*}[!ht]
	\centering{
	\hspace{-6mm}
    \subfigure[\cpd]{ \label{fig:hparams:cpd_lambda}
     \includegraphics[width=0.5\linewidth]{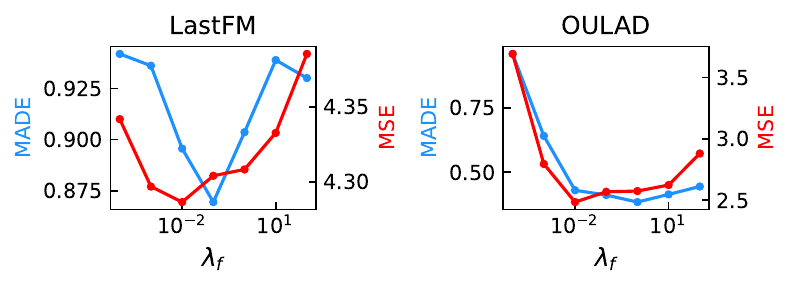}
    } \hspace{-6mm}
    \subfigure[\costco]{ \label{fig:hparams:costco_lambda}
     \includegraphics[width=0.47\linewidth]{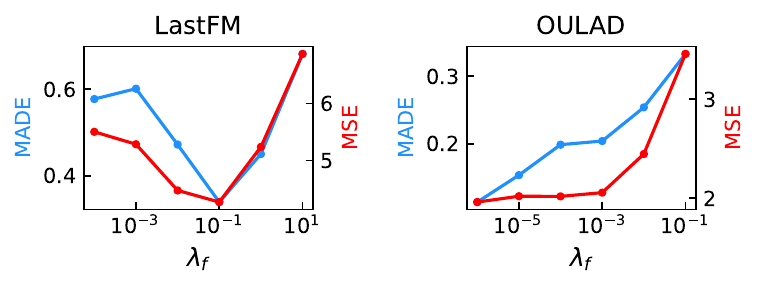}
    } \\
    \hspace{-5mm}
    \subfigure[\cpd]{ \label{fig:hparams:cpd_gamma}
     \includegraphics[width=0.50\linewidth]{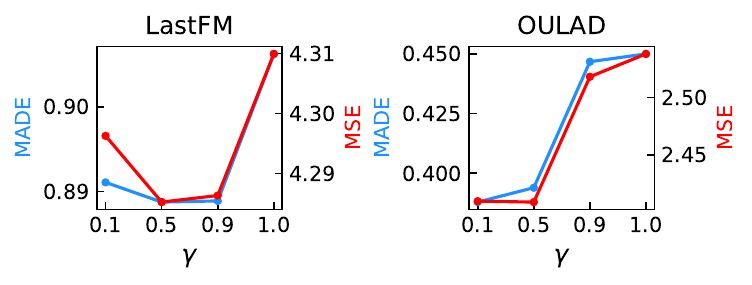}
    } \hspace{-5.5mm}
    \subfigure[\costco]{ \label{fig:hparams:costco_gamma}
     \includegraphics[width=0.49\linewidth]{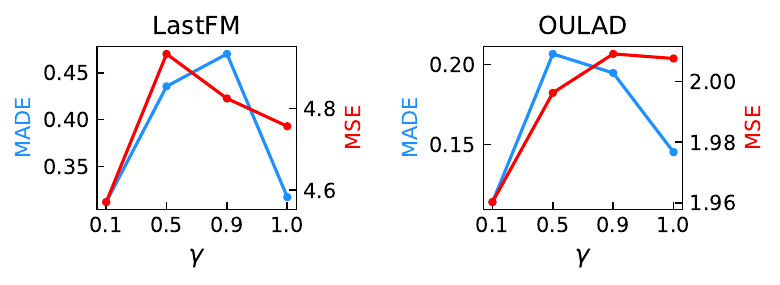}
    }
    \caption{\label{fig:hparams}
    Hyper-parameter study of $\lambda_f$ (first row) and $\gamma$ (second row) in \method with \lastfm and \oulad.}
 }
\end{figure*}

We examine the sensitivity of $\lambda_f$ for the proposed regularization and the effect of $\gamma$ in the fairness-aware graph using \lastfm and \oulad as shown in \Cref{fig:hparams}.
While fixing other hyper-parameters, 
we adjust $\lambda_f$ within the range of $[1e^{-6}, 1e^{+2}]$ according to datasets and models
and $\gamma$ from [$0.1, 0.5, 0.9, 1$].
When $\gamma = 1$, only a context feature is used to construct the $K$-NN graph
while $\gamma$ decreases, sensitive features are more incorporated to construct the $K$-NN graph.
As shown in \Cref{fig:hparams:cpd_lambda,fig:hparams:costco_lambda}, 
when $\lambda_f$ is set too large or too small, both MSE and MADE tend to increase in all tensor models.
 Specifically, \cpd shows the lowest MSE and MADE within the range of $[1e^{-2}, 1e^{+1}]$ with \oulad data, a very sparse tensor.
 This indicates that \method provides effective augmentation for sensitive entities with very sparse data when CPD struggles to capture latent patterns accurately for the sensitive entities. 
On the other hand, MSE and MADE reduce as the $\lambda_f$ decreases in \costco; 
\costco lowers MADE and MSE with a smaller $\lambda_f$ as they capture better latent patterns than CPD for sparse tensors.
\Cref{fig:hparams:cpd_gamma,fig:hparams:costco_gamma} show the lowest MADE and MSE when $\gamma$ is not equal to 1, indicating that using sensitive features to construct $K$-NN graph is effective to achieve lower MADE while not hurting MSE.

\vspace{-3mm}
\section{Conclusion}
\label{sec:conclusion}
\vspace{-3mm}
In this paper, we propose \method, a fairness-aware augmentation for sparse tensor completion.
\method constructs a fairness-aware $K$-NN graph to guide the generation of an augmented tensor, and regularizes the original entities with the augmented ones.
Our experimental results show that 
\method consistently achieves the utility and the group fairness in tensor completion
by exhibiting the best trade-off between MSE and MADE in both traditional and deep learning-based tensor models.
\method remains effective even when there is a large gap in the number of entries between the minority and majority groups.

\begin{credits}
\subsubsection{\ackname}
We would like to thank Uday Singh Saini, Rutuja Gurav, and Ryan Aschoff for valuable discussions that helped improve this work.
Research was supported by the National Science Foundation under CAREER grant no. IIS 2046086 and CREST Center for Multidisciplinary Research Excellence in CyberPhysical Infrastructure Systems (MECIS) grant no. 2112650,  by the Agriculture and Food Research Initiative Competitive Grant no. 2020-69012-31914 from the USDA National Institute of Food and Agriculture, and by the University Transportation Center for Railway Safety (UTCRS) at UTRGV through the USDOT UTC Program under Grant No. 69A3552348340.
Jun-Gi Jang was supported by Basic Science Research Program through the National Research Foundation of Korea (NRF) funded by the Ministry of Education (RS-2023-00238596).
%
\end{credits}

\bibliographystyle{splncs04}
\bibliography{bib/dawon}

\end{document}